# Robust speech recognition using consensus function based on multi-layer networks

Rimah Amami
Department of Electrical Engineering
National School of Engineering
Tunis, Tunisia
Email: rimah.amami@gmail.com

Ghaith Manita
Department of Software Engineering,
Higher Institute of Computer Science
Tunis, Tunisia
Email: gaith.manita@gmail.com

Abir Smiti
LARODEC, University of Tunis
Institut Supérieur de Gestion,
Tunis, Tunisia
Email: smiti.abir@gmail.com

*Abstract* — **The clustering ensembles mingle numerous partitions of a specified data into a single clustering solution. Clustering ensemble has emerged as a potent approach for ameliorating both the forcefulness and the stability of unsupervised classification results. One of the major problems in clustering ensembles is to find the best consensus function. Finding final partition from different clustering results requires skillfulness and robustness of the classification algorithm. In addition, the major problem with the consensus function is its sensitivity to the used data sets quality. This limitation is due to the existence of noisy, silence or redundant data. This paper proposes a novel consensus function of cluster ensembles based on Multilayer networks technique and a maintenance database method. This maintenance database approach is used in order to handle any given noisy speech and, thus, to guarantee the quality of databases. This can generates good results and efficient data partitions. To show its effectiveness, we support our strategy with empirical evaluation using distorted speech from Aurora speech databases.**

## I. INTRODUCTION

The objective of cluster analysis is to assemble a team of related instances into sets that in some sense fit in together because of related attributes. Almost all of the well-known clustering techniques face many challenges and difficulties such as the necessitated number of clusters for the forecast which is hard to set. Additionally, the preference of the adequate similarity measure and the identification of the noisy points are crucial tasks for the systems robustness. In order to solve those issues, the idea of combining two or more clustering techniques produced by multiple algorithms has recently been proposed [1], [2].

Cluster ensemble has attested to be an excellent option when facing cluster analysis problems [3]. It consists of generating a set of clusterings from the same dataset and combining them into a ultimate clustering. Such ensembles can be achieved, for example, by changeable the (hyper) parameters of a "base" clustering algorithm, by resembling or reweighing the set of objects, or by employing several different base clusters.

The purpose of this amalgamation process is to progress the value of individual data clusterings. It can provide more robust and stable solutions by making use of the consensus across multiple clustering results, while averaging out emergent spurious structures that arise due to the various biases to which each participating algorithm is tuned, or to the variance induced by different data samples [4].

The consensus function is applied to combine results of the different clustering techniques. The result of different partition sets is combined to attend more representative one. Thus, there are two fundamental components which are the mechanism used to generate initial partitions and the consensus function used to combine these partitions into a final one.

The Cluster ensembles is applied to a wide range of applications. It is also used as a classifier for a static machine learning data sets, where it was shown that the algorithm can process data classification. Nevertheless, it suffers from some shortcomings as it is sensitive to the data type [5].

The main reason of this drawback is related to the quality of the database which contains disagreeable objects such as noisy, inconsistent or redundant instances. This situation may occur due to several reasons like the existence of many sources of anomaly detection that makes the database reach of noises. Therefore, it affects negatively its classification results. To handle this drawback, maintaining the database, as a first step, becomes necessary. Hence, the first objective of our paper is to use a maintenance policy.

Actually, various maintenance policies have been proposed in literature [6]. Most of these methods can offer an acceptable database size reduction and satisfying classification accuracy; despite some of them are expensive to run and neglect the importance of the noisy instances elimination. Based on our first objective, we choose to apply the maintaining method which is based on the Clustering technique called SOFT-DBSCAN [7] for the given database. Our maintenance approach is originally applied in Case Based Reasoning technique and it seems appropriate to our case since it is characterized by its capability of removing noisy an redundant instances as well its ability to improve the classification accuracy and offering a reasonable execution time.

Thus, we spotlight the dilemma of combining multiple partitioning of a set of objects into a single consolidated clustering. For that, we propose in this paper a novel consensus functions of cluster ensembles based on Multilayer networks technique and a maintenance database method. This maintenance database approach is used in order to handle any given noisy speech and, thus, to guarantee the quality of databases. Our new method of cluster ensembles named RCFM - Robust Consensus Function based on Multi layer networks. This can generates good results and efficient data partitions.

To show its effectiveness, we support our strategy with empirical evaluation using distorted speech from Aurora speech database.

The rest of this paper is organized as follows: in the next Section we introduce a brief overview of the cluster ensemble methods. Section 3 describes our work proposed approach and discusses their benefits. Experimental setup and results are given in section 4. The paper concludes in Section 5

## II. RELATED WORKS

Various divisions are classically produced by using different clustering algorithms, or by applying a single algorithm with different parameter settings, possibly in combination with data or feature sampling. A cluster ensemble technique is characterized by two components: the mechanism to generate diverse partitions, and the consensus function to combine the input partitions into a final clustering.

The first component involves several methods which generates different clustering results from data sets. This process may be done by applying a method such as using one algorithm with different built-in initialization and parameters, selecting different subsets of data points, projecting data into different subspaces, or projecting various clustering algorithms.

The purpose of the second component is to merge initial clusterers into a final one using mathematical functions andalgorithms called consensus functions. The final clusterer provided by those functions is constructed by six types of approaches: (1) hypergraph methods, (2) voting approaches, (3) information theoretic methods, (4) co-association-based methods, (5) mixture models, and (6) evolutionary algorithms.

In this paper, we decide to explore the voting approach, in order to generate the final clusterer. The vote strategys objective is to solve the correspondence problem between the labels of known and derived clusters by finding the corresponding cluster labels among multiple partitions, then obtain the consensus partition through a voting process [8].

The main idea is to change the cluster labels to match up with the best arrangement between the labels of two partitions. Then, the generated partitions have to be relabeled according to a predefined partition selected from either the ensemble or a new clustering of the data sets known as the reference partition. The research works carried out by [9] [10] contribute to study the number of clusters where each given partition is the same as in the target partition. In this context, we present a new approach that can handle the problem of combining multiple clusterings, propose a suitable objective function for determining a single consensus clustering, and explore the feasibility of directly optimizing this objective function using multi-layer networks approaches. Before presenting the steps of our method, let us, first, make a brief induction to the multi-layer networks since we deal with it. Moreover, the consensus functions have greater difficulty with the problem of the combinatorial optimization problem. In this context, we propose a new approach which handles the problem of combining multiple clusterings.

## III. ROBUST CONSENSUS FUNCTION BASED ON MULTI- LAYER NETWORKS

According to literature, the consensus function suffers from some shortcomings as it is sensitive to the type of data. This is due to an environment characterized by the existence of redundant and noisy instances in the given database. Hence, in this paper, we propose a new method of cluster ensembles named RCFM - Robust Consensus Function based on Mult layer networks. Our method takes into account the fact of alleviating the maintaining the particular database by eliminating its "useless" objects. Our RCFM steps are described in the following Subsections.

### A. Maintaining the database

As stated previously, a given database contains disagreeable objects especially noisy and redundant instances, in the sense that it affects negatively the quality of the classification results and the results of the clustering ensembles. To guarantee the consensus functions performance, the maintenance of the database becomes essentially.

Thus, we decide to apply the Clustering method SOFTDBSCAN [7], [11] as a first step in our new RCFM. In fact, applying SOFT-DBSCAN in the database aims at eliminating its noisy and redundant instances.

Before starting to explain our maintenance strategy, it is worth answering these question: which instances should be kept in the database, which ones should be eliminated, and why? In order to have a good database quality, we should keep instances whose deletion directly reduce the accuracy of the classification.

We need first to create multiple, groups from the database that are located on different sites. Each group contains points that are closely related to each other. In that way, we can define the accuracy group. This can be done by a clustering technique. For each cluster, the point which expected to be noisy, which in their turns decrease the quality of the group, are removed and the rest of cases is kept. Therefore, we obtain a new small database with high competence. The basic process of our proposed maintaining method is based on a clustering method. Actually, clustering solves our noisy problems. The use of clustering in database offers many positive points: It creates groups which are easy to treat, it facilitates the base maintainer since case bases are often large and unstructured, it allows to define and calculate accuracy group to preserve much competence. Among several clustering approaches, we should, ideally, use a clustering method that discovers structure in such data sets and has the maximum of the following main properties:

- It has the capability of handling noisy cases: Noises are a distortion of a value or the addition of the spurious object. The noisy instances are disagreeable objects, they can dramatically slow the classification accuracy. As a result, the datas performance will be decreased and the the quality will be reduced. Hence, they should be eliminated.
- It scales up for large database.
- It can create clusters with different shapes
- It allows the elements to have a degree of membership for each cluster, because we believe that one case can belong to more than one group.

To overcome all these conditions, we use the fuzzy clustering method named "SOFT DBSCAN" proposed in [7]. This technique is an appropriate clustering method for our maintenance technique because it has a number of good aspects: it can create regions which may have an arbitrary shape and the points inside a region may be arbitrarily distributed, it can detect points expected to be noises and it is able to assign one data point into more than one cluster by affecting to each observation a "degree of membership" to each of the classes in a way that is consistent with the distribution of the data.

We can resume the basic steps of this clustering technique as follows:

**Algorithm 1** Basic Soft DBSCAN Algorithm
1: Begin
2: m: weighting exponent ($m > 2$)
3: $\xi$: tolerance level
4: Run DBSCAN and find:
   x = number of noises
   k = number of clusters
5: $c \leftarrow x + k$
6: Create the initial fuzzy partition:
   if $x_i \in c_j$ then $u_{ij} \leftarrow 1$
   Else $u_{ij} \leftarrow 0$
7: $t \leftarrow 0$
8: Repeat
   Update $U_t$ as following: $cr\mu_{ik} = [\sum_{j=1}^{c}(\frac{MD_{ik}}{MD_{jk}})^{\frac{2}{m-1}}]^{-1}$
   Where $MD_{ik}$ is the Mahalanobis distance between $x_k$ and $v_k$
   Calculate $v_t$ as following
   $v_i = \frac{1}{\sum_{k=1}^{n} \mu_{ik}^m} \sum_{k=1}^{n} \mu_{ik}^m x_{ik}$   i= 1, 2,...,c
9: Until $\|U_t - U_{t-1}\| \leq \xi$
10: $(U, v) \leftarrow (U_t, v_t)$
11: noisy points = $\{x_{ij} | c_j = x_{ij}\}$
12: End

Consequently, the result of applying the maintaining method is the generation of a new reduced database lacking noisy and redundant objects while preserving nearly the same performance of the original data set. Hence, our RCFM can treat the new database easily and it guaranties better classification results and facilitate the role of the clustering ensembles.

*B. MLNCF -Multi Layer Networks for Consensus Function*

Once the given database is maintained using our policy described in the first step, we build a suitable objective function for determining a single consensus clustering, denoted MLNCF -Multi Layer Networks for Consensus Function- and we explore the feasibility of directly optimizing this objective function using greedy approaches Multi Layer Networks (MLN). We can describe the basic process of our strategy in the following steps:

For the first step, we run different clustering methods on our given datasets S. $C_k$ denotes the result of this step which consists on K independent clusters. The data are partitioned on K clusters.

For the second step, we trait every cluster $C_i$ apart. We apply Multi layer Networks (MLN) [12]. Actually, Network theory is an important tool for describing and analyzing complex systems. As research on complex systems has matured, it has become increasingly essential to move beyond simple graphs and to investigate more complicated but more realistic frameworks. MLN crack the classification dilemma for non linear sets by using hidden layers, whose neurons are not straightforwardly attached to the output. The extra hidden layers can be deduced geometrically as additional hyperplanes, which improve the division capacity of the network.

For that, we use an unique architecture network for all the learning process. Our approach computes the weights matrix using gradient descent (GD). Actually, the GD [13] is able to modify weights in order to decrease the error function (E) over all training samples of the networks, by adapting weights in MLN:

$$E(t) = \frac{1}{2} \sum_{h=1}^{} (d_h(t) - Y_h(t))^2 \qquad (1)$$

Where $d_h$ is the regret value for dimension h.

In our case, the network is erected from the input layer, hidden layers and the output layer. For each training sample, the neurons in the same layer are autonomous, However, the neurons in adjacent layers are totally connected with the weight W generated by GD. We supposed that for the input X and the output Y is as follows:

$$Y = F(X, W, \Theta) \qquad (2)$$

Where $\Theta$ is the bias. The network error is decreased by varying W.

In the last step, we combine the computing $W_j$ of all clusters into a single matrix $W_f$ using samples arithmetic means. After that, MLN is executed on $W_f$ to generate the optimal and final cluster.

We can resume these steps as follows [14], [15]:

---
**Algorithm 2** MLNCF -Multi Layer Networks for Consensus Function
---
1: Begin
2: **Input**: dataset S=(N*M)
3: Applying different clustering methods on S
4: Construct diverse clusterers $C_i$ where $i \in 1, K$ and k is the number of clusters.
5: For each cluster $C_i$ Generating learning model using MLN in order to Compute the weights $W_{ij}$ matrix.
6: Combining the $W_{ij}$ matrix of all clusters into one matrix $W_f$ using arithmetic means.
7: Applying MLN on S using the result weight matrix $W_f$ to generate the final cluster.
8: **Output**: The final cluster $C_f$
9: End
---

## IV. EXPERIMENTAL ANALYSIS

We have developed our programs in Matlab V7.1 for the evaluation of our RCFM method. To show its effectiveness, we support our strategy with empirical evaluation using distorted speech from Aurora speech database. The next subsection, we will describe this base.

### A. A. Speech system description

Automatic Speech recognition (ASR) applications are becoming more and more useful nowadays and widely used in many real-world problems.

In this paper, we study continuous speech recognition based on TIdigits units found by an unsupervised splitting algorithm. The database Aurora 2 is consisting of connected digits task spoken by Native American English speakers. Moreover, this database is designed to evaluate the performance of speech recognition algorithms in noisy conditions. Thus, a selection of several real-world background noise have been added to the speech at different signal-to-noise ratios (SNRs) [16]. The architecture of the speech recognition systemis illustrated in figure 1.

At first step, the input noisy speech analog signal is preprocessed which includes the steps of sampling, filtering, add window, etc. The Mel-Frequency Cepstral Coefficients (MFCC) are popular features representations. Indeed, those acoustic coefficients are the most commonly used for ASR systems.

The main idea of this algorithm consider that the MFCC are the cepstral coefficients calculated from the mel-frequency warped Fourier transform representation of the log magnitude spectrum.

Including the temporal cepstral derivative aim to improve the performance of speech recognition system. Those coefficients have shown a determinant capability to capture the transitional characteristics of the speech signal that can contribute to ameliorate the recognition task [17].

At the second step, the noisy speech signals are scanned by our maintaining algorithm, described in the previous Section, in order to clean up the database from the noise, useless data and/ or silence period. Then, the recognition system proceeds with the training step by applying a several clustering algorithms in aim to find out the most similar groups from the TIdigit data. The model outputted from each algorithm will be combined using a consensus function. The resulted model will be used to recognize new digit samples from the test database.

The experiments results will show, in the next section, the influence of the use of distorted data by adding real noise effects on the global performance of the speech recognition system. Then, we will compare the results after maintaining the database to eliminate and/ or reduce the noise.

### B. Results and Discussion

The Aurora 2 database is used for experimentation. There are the recordings of male and female English American adults speaking isolated digits. Each speaker pronounces sequences of up to 7 digits sampled at a rate of 8kHz. A feature vector of 39-dimensional MFCC coefficients was obtained and used. The performance of test set is computed for the SNR equal to 0 dB. It must be pointed out that the selected noise signals have been recorded at different places:

- Street
- Train station
- Airport
- Suburban train
- Restaurant
- babble
- Exhibition hall
- Car

In this section, we present and compare the recognition results when applying three training data:

1) Training on noisy TIdigit data before using DBSCAN and consensus function (K-means, PAM, Fuzzy Cmeans).

2) Training on noisy TIdigit data using consensus function (MLCF).

3) Training on clean TIdigit data after using DBSCAN and consensus function (RCFM).

The word recognition accuracy is shown in table 1 for data sets with noisy background and, first, without applying a consensus function and with a consensus function. The results in table 1 demonstrated that the recognition system showed poor performance the all selected noises conditions. Meanwhile, The application of a consensus function permit to improves the overall performance of the word recognition.

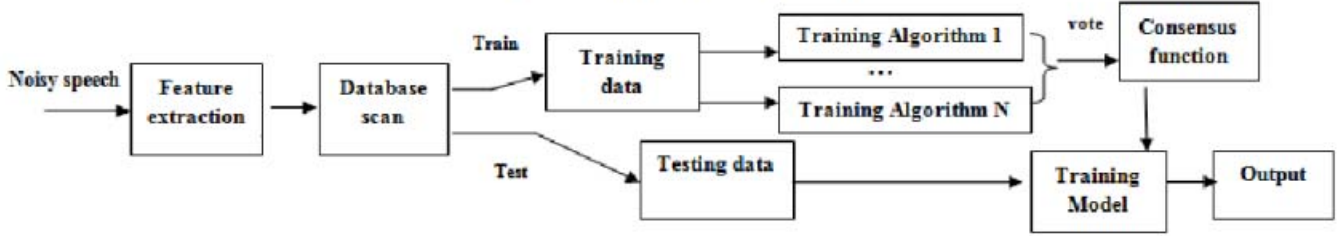

Figure 1. Speech system architecture

TABLE I. TIdigit recognition accuracy as percentage for test set with noisy training data

| Methods | Subway | Restaurant | Babble | Street | Car | Airport | Exhibition | Train-station |
|---|---|---|---|---|---|---|---|---|
| K-means | 57.66 | 50.76 | 44.57 | 52.44 | 47.41 | 49.87 | 55.54 | 46.23 |
| PAM | 54.17 | 49.36 | 42.32 | 50.78 | 47.21 | 48.86 | 52.43 | 44.75 |
| Fuzzy C-means | 59.72 | 52.38 | 48.55 | 54.89 | 50.24 | 51.42 | 57.31 | 49.77 |
| MLCF | 81.02 | 77.63 | 70.12 | 78.22 | 73.15 | 75.64 | 72.43 | 79.89 |

TABLE II. TIdigit recognition accuracy as percentage for test set with clean training data and consensus function

| Methods | Subway | Restaurant | Babble | Street | Car | Airport | Exhibition | Train-station |
|---|---|---|---|---|---|---|---|---|
| K-Means | 61.09 | 59.42 | 56.07 | 59.31 | 58.21 | 58.71 | 60.32 | 55.23 |
| PAM | 59.35 | 51.87 | 46.47 | 53.89 | 50.76 | 51.04 | 57.02 | 45.88 |
| Fuzzy C-means | 64.12 | 61.04 | 57.23 | 62.11 | 58.75 | 59.33 | 66.78 | 58.01 |
| RCFM | 95.33 | 92.21 | 87.43 | 93.74 | 89.43 | 90.11 | 94.51 | 89.01 |

The results for the test set in the table 2 show a improvements of the TIdigit recognition after maintaining the database with the DBSCAN algorithm for all the clustering methods. This database scan allows the reduction of the noise which distorted the speech signals leads to better results. Moreover, the use of a consensus function allows to enhance the performance of the word recognition.

The word recognition accuracy is shown in table 2 for data sets when using clean data and applying a consensus function to form a combined model from the different cluster ensembles methods.

The results for the test set show significant improvements for all the recoding conditions. In fact, the use of both consensus function and an algorithm to reduce the noise and clean the database leads to increase the performance of the recognition system.

The best performance is produced with the data with the subway noise within 95.33% applying RCFM approach. In final, the comparison between the results presented in table 1 et and table 2 shows that the performance is worse in noise conditions. This result can be explained by the fact that the different clustering ensembles methods cannot deals with the noisy data.

V. CONCLUSION

In this paper, we have developed a modified version of a previous consensus function method. The called RCFM - Robust Consensus Function is based on Multi layer networks. Our new method aims to maintaining the speech databases, in order to generate a new database lacking noisy and redundant elements while preserving nearly the same performance of the original data sets. The proposed technique lead to better results in terms of speech classification accuracy.

We conducted experiments to show how cluster ensembles can be used to introduce robustness clustering algorithms, and dramatically improve groups of subspace clusterings for our noisy data sets from Aurora speech corpus.

As future work, we intend to further explore this new instantiation of our RCFM by introducing soft computing algorithms in uncertain context methods in order to check the reliability of our cluster ensemble technique.

REFERENCES

[1] M. Bajaj and A. Nain, "Improved result by combining two clustering techniques," International Journal of Advanced Research in Computer Science and Software Engineering, pp. 862 – 867, 2014.
[2] A. Fred and A. Jain, "Combining multiple clusterings using evidence accumulation," Pattern Analysis and Machine Intelligence, pp. 835–850, 2005.
[3] A. Strehl, "Cluster ensembles-a knowledge reuse framework for combining multiple partitions," Journal of Machine Learning Research, pp. 583–617, 2002.
[4] C. Domeniconi and M. Al-Razgan, "Weighted cluster ensembles: Methods and analysis," ACM Trans. Knowl. Discov. Data, vol. 2, no. 4, pp. 17:1–17:40, Jan. 2009.

[5] S. Hadjitodorov and L. I. Kuncheva, "Selecting diversifying heuristics for cluster ensembles," Multiple Classifier Systems, Lecture Notes in Computer Science, pp. 200–209, 2009.

[6] A. Smiti and Z. Elouedi, "Article: Overview of maintenance for case based reasoning systems," International Journal of Computer Applications, vol. 32, no. 2, pp. 49–56, October 2011, published by Foundation of Computer Science, New York, USA.

[7] ——, "Soft dbscan: Improving dbscan clustering method using fuzzy set theory," The 6th International Conference on Human System Interaction HSI2013, Spot, Poland, pp. 380 – 385, 2013.

[8] S. D. E. Dimitriadou and A. Weingessel, "An examination of indexes for determining the number of clusters in binary data sets," Psychometrika Society, Springer, pp. 137–159, 2002.

[9] S. Dudoit and J. Fridlyand, "Bagging to improve the accuracy of aclustering procedure." Bioinformatics, vol. 19, no. 9, pp. 1090–1099, 2003.

[10] B. Fischer and J. M. Buhmann, "Path-based clustering for grouping of smooth curves and texture segmentation," IEEE TRANSACTIONS ON PATTERN ANALYSIS AND MACHINE INTELLIGENCE, vol. 25, no. 4, pp. 513–518, 2003.

[11] A. Smiti and Z. Elouedi, "Dbscan-gm: An improved clustering method based on gaussian means and dbscan techniques," in International Conference on Intelligent Engineering Systems (INES), Portugal. IEEE Computer Society, June 2012, pp. 573–578.

[12] M. Kivel¨a, A. Arenas, M. Barthelemy, J. P. Gleeson, Y. Moreno, and M. A. Porter, "Multilayer networks," CoRR, vol. abs/1309.7233, 2013.

[13] J. A. Snyman, Practical Mathematical Optimization: An Introduction to Basic Optimization Theory and Classical and New Gradient-Based Algorithms, ser. Applied Optimization, Vol. 97. Springer-Verlag New York, Inc., 2005.

[14] G. Manita, "Consensus function based on multi-layer networks technique," in Human System Interaction (HSI), 2013 The 6th International Conference on, June 2013, pp. 252–256.

[15] R. K. G. Manita and M. Limam, "Consensus functions for cluster ensembles," Applied Artificial Intelligence, vol. 26, no. 6, pp. 598–614, 2012.

[16] D. P. H. G. Hirsch, "The aurora experimental framework for the performance evaluation of speech recognition systems under noisy conditions," ISCA ITRW ASR2000, pp. 18–20, 2000.

[17] S. Davis and P. Mermelstein, "Comparison of parametric representations for monosyllabic word recognition in continuously spoken sentences," International conference on acoustic speech and signal processing, vol. 28, pp. 357–366, 1980